\documentclass{bmvc2k}

\title{Light-Weight RetinaNet for Object Detection}

\runninghead{Y. Li and F. Ren}{Light-Weight RetinaNet for Object Detection}

\addauthor{Yixing Li}{yixingli@asu.edu}{1}
\addauthor{Fengbo Ren}{renfengbo@asu.edu}{1}

\addinstitution{
 School of Computing, Informatics, and Decision Systems Engineering\\
 Arizona State University\\
}

\begin{document}

\maketitle

\begin{abstract}
Object detection has gained great progress driven by the development of deep learning. 
Compared with a widely studied task -- classification, generally speaking, object detection even need one or two orders of magnitude more FLOPs (floating point operations) in processing the inference task. To enable a practical application, it is essential to explore effective runtime and accuracy trade-off scheme. Recently, a growing number of studies are intended for object detection on resource constraint devices, such as YOLOv1, YOLOv2, SSD, MobileNetv2-SSDLite \cite{yolov1,yolov2,ssd}, whose accuracy on COCO test-dev \cite{cocodataset} detection results are yield to mAP around 22-25\% (mAP-20-tier). On the contrary, very few studies discuss the computation and accuracy trade-off scheme for mAP-30-tier detection networks. 
In this paper, we illustrate the insights of why RetinaNet gives effective computation and accuracy trade-off for object detection and how to build a light-weight RetinaNet. We propose to only reduce FLOPs in computational intensive layers and keep other layer the same. Compared with most common way -- input image scaling for FLOPs-accuracy trade-off, the proposed solution shows a constantly better FLOPs-mAP trade-off line. Quantitatively, the proposed method result in 0.1\% mAP improvement at 1.15x FLOPs reduction and 0.3\% mAP improvement at 1.8x FLOPs reduction.

\end{abstract}

\section{Introduction}
\label{sec:intro}
Object detection serves as an important role in computer vision-based tasks \cite{fasterrcnn,yolov1,retinanet}. It is the key module in face detection, tracking objects, video surveillance, pedestrian detection etc \cite{deepface,wang2013intelligent}. With the recent development of deep learning, it boosts the performance of object detection tasks. However, regarding the computational complexity (in terms of FLOPs), the detection network can possibly consume three orders of magnitude more FLOPs than a classification network, which makes it much more difficult to move towards low-latency inference. 

Recently, there have been growing number of studies investigating into detection on resource constraint devices, such as mobile platforms. As the main concern of resource constraint devices is the memory consumption, the existing solutions, such as YOLOv1, YOLOv2, SSD, MobileNetv2-SSDLite \cite{yolov1,yolov2,ssd} have pushed it hard to reduce the memory consumption by trading off their accuracy performance. Their accuracy on large dataset -- COCO test-dev 2017 \cite{cocodataset} detection results are yield to mAP of 22-25\%. Here, we use the mAP as the indicator to categorize these solutions as mAP-20-tier. On the other side, in the mAP-30-tier, popular solutions can be listed as Faster R-CNN, RetinaNet, YOLOv3 \cite{fasterrcnn,retinanet,yolov3} and their variants. As these solutions are commonly deployed on mid- or high-end GPUs or FPGAs, the memory resource is probably enough for preloading the weights. In addtion, \cite{speedandaccu} also verifies the linear relation between number FLOPs and inference runtime for the same kind of network. By applying Faster R-CNN, RetinaNet and YOLOv3 \cite{fasterrcnn,retinanet,yolov3} on the same task for COCO detection dataset, which takes an input image around 600x600-800x800, the mAP will hit in the range of 33\%-36\%. However, the FLOPs of Faster R-CNN \cite{fasterrcnn} is around 850 GFLOPs (gigaFLOPs), which is at lease 5x more than that of RetinaNet and YOLOv3 \cite{yolov3}. Apparently, Faster R-CNN is not competitive in the computational efficiency. From YOLOv2 \cite{yolov2} to YOLOv3 \cite{yolov3}, it is interesting that the authors have aggressively increased the number of FLOPs from 30 to 140 GFLOPs to gain mAP improvement from 21\% to 33\%. Even with that, its mAP is 2.5\% lower than RetinaNet with 150 GFLOPs. This observation inspires us to take the RetinaNet as the baseline to explore a more light-weight version. 

There are two common methods to reduce the FLOPs in a detection network. One way is to switch to another backbone, while the other is reducing the input image size. The first one definitely results in noticeable accuracy drop if one switches from one of the ResNet backbones \cite{resnet} to the other. Typically it won't be consider as a good accuracy-FLOPs trade-off scheme with small variation. With regard to reduce the input image size, it is an intuitive way to reduce the FLOPs. However, the accuracy-FLOPs trade-off line shows degradation in an exponentially trend \cite{speedandaccu}. There is an opportunity to find a more linear degradation trendancy line for better accuracy-FLOPs trade-off. We propose to only replace the certain branches/layers of the detection network with light-weight architecture and keep the rest of the network unchanged. For the RetinaNet, the heaviest branch is the succeeding layers of the finest FPN (P3 in Fig. 1), which takes up 48\% of the total FLOPs. We propose different light-weight architecture variants.  More importantly, the proposed method can also be applied to other blockwise-FLOPs-imbalance detection networks. 

The contributions of this paper can be summarized as follows: (1) We proposed only to reduce the heaviest bottleneck layer for the light-weight RetinaNet mAP-FLOPs trade-off. (2) The proposed solution shows a constantly better mAP-FLOPs trade-off line in a linear degradation trend, while the input image scaling method degrades in a more exponentially trend. (3) Quantitatively, the proposed method result in 0.1\% mAP improvement at 1.15x FLOPs reduction and 0.1\% mAP improvement at 1.15x FLOPs reduction and 0.3\% mAP improvement at 1.8x FLOPs reduction.

\section{Related Works}
\subsection{High-end object detection networks (mAP-30-tier)}
Faster RCNN \cite{fasterrcnn} is an advanced architecture, which boosts both the accuracy and runtime performance from R-CNN and Fast R-CNN \cite{rcnn,fastrcnn}. 

The main body of Faster RCNN \cite{fasterrcnn} is composed of three parts -- Feature Network, Region Proposal Network (RPN) and Detection Network. As Faster RCNN \cite{fasterrcnn} replaces the selective search (used by Fast R-CNN) with RPN, it significantly reduces the runtime of gernerate the region proposals. However, in the Faster R-CNN inference stage, there are still around 256-1000 boxes will feed into the detection network, which is really expensive to process this large batch of data. As for a Faster RCNN to process COCO dataset detection task with Inception-ResNetV2 \cite{inceptionv2} backbone, the total numbers of FLOPs can comes up to over 800 GFLOPs.

Compared with Faster RCNN \cite{fasterrcnn}, the RetinaNet \cite{retinanet} targets a simpler design for gaining speedup. A feature pyramid network (FPN) \cite{fpn} is attached to its backbone to generate multi-scale pyramid features. Then, pyramid features go into classification and regression branches, whose weights can be shared across different levels of the FPN. The focal loss is applied to compensate the accuracy drop, which makes its accuracy performance to be comparable with the Faster RCNN.

\begin{figure*}
\begin{center}
\includegraphics[width=11cm]{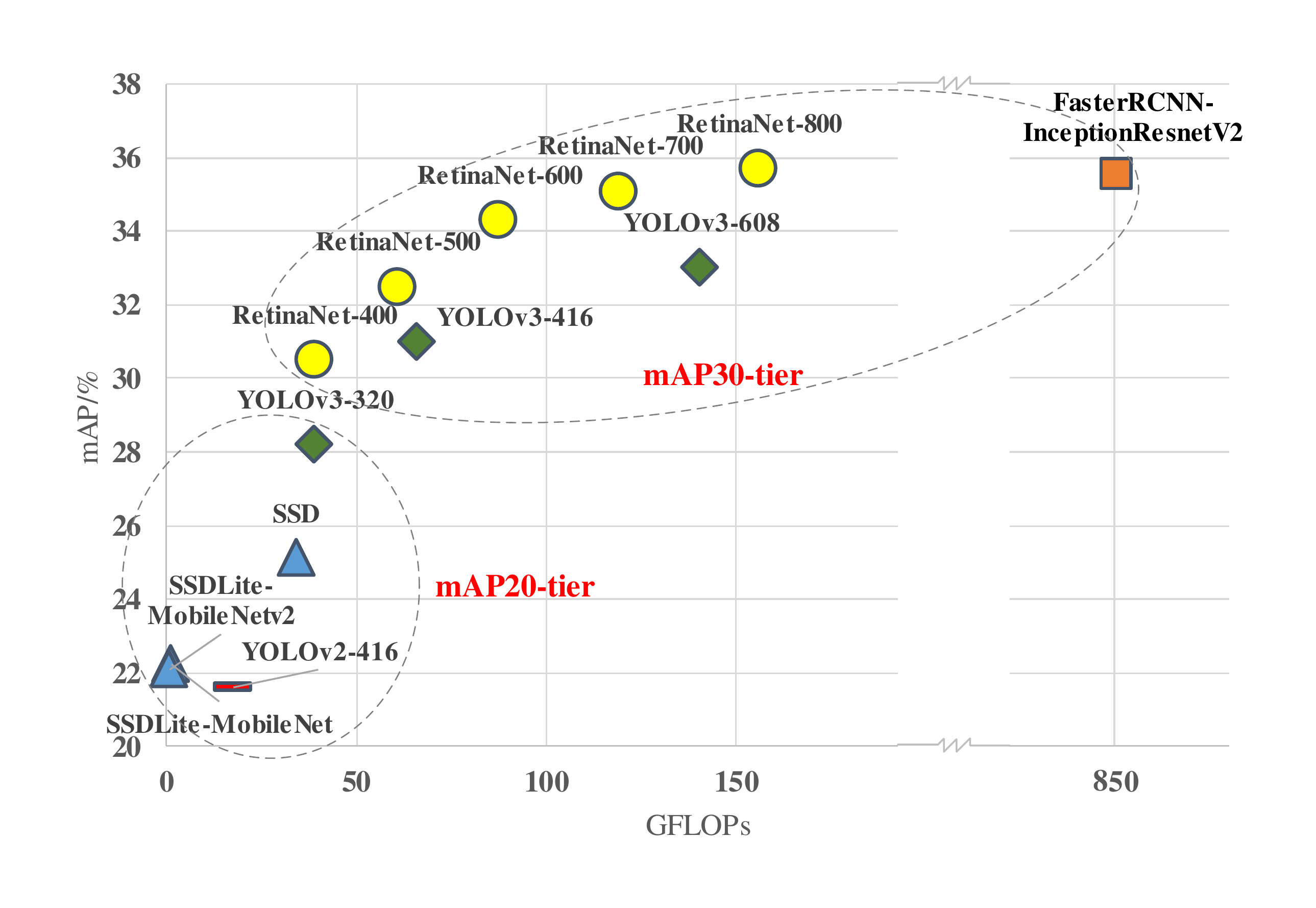}
\end{center}
   \caption{Example of a short caption, which should be centered.}
\label{fig:short}
\end{figure*}

\subsection{Light-weight object detection networks (mAP-20-tier)}
The YOLO network family are among the most popular ones. The most distinguished feature of YOLO is their predefined grid cell. The input image can be cut into SxS grid cells, and each cell only predicts one object. On the good side, this idea apparently helps to reduce the computation complexity. However, in the meantime, it increases the chances of undetected objects and has relative bad performance on detecting small objects. The YOLOv1 \cite{yolov1} is only evaluated on relative small datasets (PASCAL VOC), with the main contribution of enabling real-time inference. From YOLOv2 \cite{yolov2} to YOLOv3 \cite{yolov3}, the mAP performance results on COCO test-dev2015 dataset is boosted from 21.6\% to 33.0\%. It's worth noting that the accuracy gain of YOLOv3 comes along with the FLOPs increment from 63 GFLOPs to 141 GFLOPs. Interestingly, YOLOv3 \cite{yolov3} should not be categorized as a light-weight one anymore, as the FLOPs counts and mAP is closed to those of RetinaNet-ResNet50-FPN (156 GFLOPs and mAP = 35.7\%). We also include other light-weight objection detection network such as SSD and SSDLite \cite{ssd} in Fig.1 for providing an overview of mAP-20-tier detection networks. 

As the RetinaNet can win over YOLOv3 on both FLOPs and mAP, this inspire us to take the RetinaNet as the baseline design to explore a better scheme for accuracy and FLOPs tradeoff for the high-end detection tasks.

\section{Light-weight RetinaNet}
In Section 2, we have explained why we think the RetinaNet network architecture has the potential to be tailored for a better accuracy and FLOPs traderoff. Here, in this Section, first we will further analyze the RetinaNet network with a focus of the distribution of the number of floating-point operations (FLOPs) across different layers in Section 3.1. Then Section 3.2 illustrates the scheme to help RetinaNet to lose weight. 
\subsection{RetinaNet Primer}
\begin{figure*}[!t]
\begin{center}
\includegraphics[width=12cm]{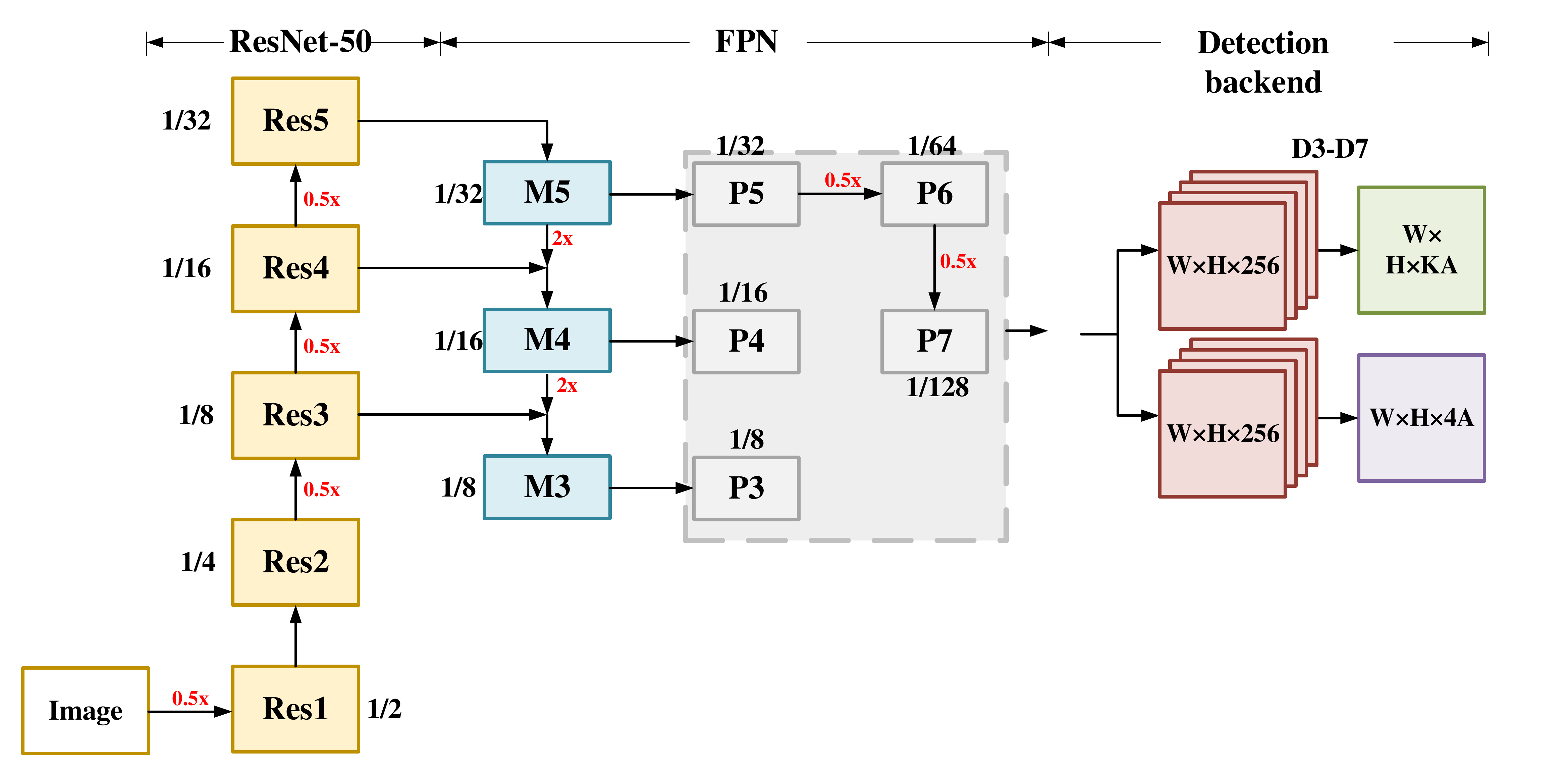}
\end{center}
   \caption{RetinaNet (ResNet50-FPN-800x800) network architecture.}
\label{fig:short}
\end{figure*}

The RetinaNet architecture is composed of three parts -- a backbone, a feature pyramid network (FPN) \cite{fpn} and a detection back-end as shown in Fig. 2. The image will first be processed by the backbone, which usually is the ResNet Architecture. Here, it is worthy to notify that although MobileNet's performance \cite{mobilenet} is on a bar with ResNet in the classification tasks, it is not true that MobileNet \cite{mobilenet} can be an equivalence substitution for the ResNet. From both \cite{speedandaccu} and our observation, use MobileNet \cite{mobilenet} as the backbone for detection task will suffer with much more accuracy drop than it does for the classification. One main reason is that the confidence scores of a MobileNet-based backbone is reduced by trading off with lower computation costs. Therefore, it wouldn't be a desirable choice of the backbone for high precision object detection networks. The backbone together with the succeeding FPN forms an encoder-decoder-like network. The benefit of the FPN is that it merges the features of consecutive layers from the coarsest to the finest level, which effectively propagate the features in different level and different scale to the succeeding layer. After then, the multi-scale pyramid features (P3-P7) will feed into the back-end where two detection branches used for bounding box regression and object classification. Note that, for the detection and bounding box branches do not share weights. While the weights of each branch are shared across pyramid features (P3-P7).

\begin{figure*}[!t]
\begin{center}
\includegraphics[width=10cm]{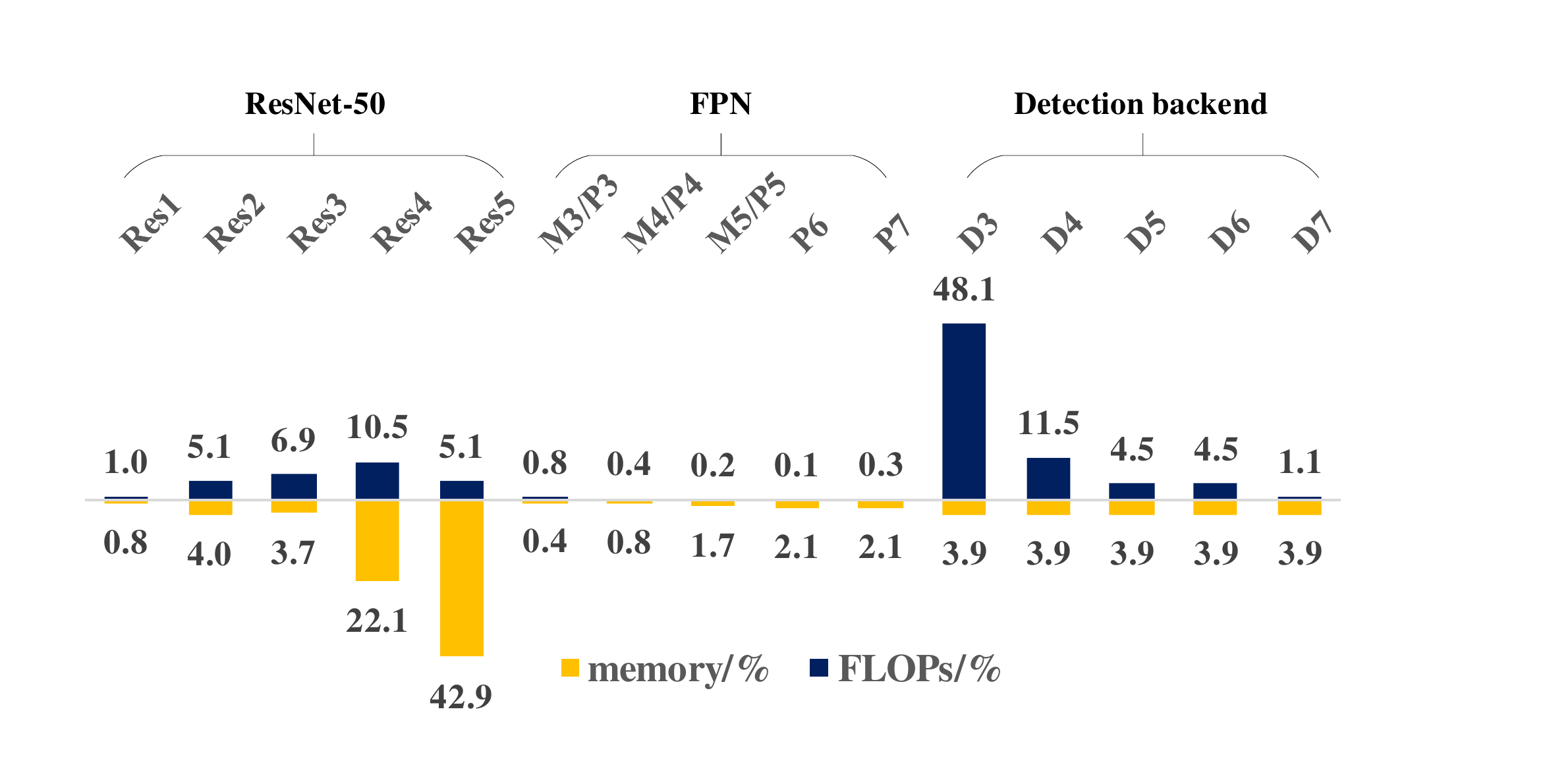}
\end{center}
   \caption{The FLOPs and memory (parameter) distribution of RetinaNet across different blocks.}
\label{fig:short}
\end{figure*}

The FLOPs distribution of RetinaNet across different blocks is shown in Fig. 3. Here, each block is corresponding to the same block in Fig. 2. For the detection backend D3-D7, they refer to succeeding layers of P3-P7, respectively. As in the original design, D3-D7 share the same weight parameters, we just show the average memory cost of D3-D7 in Fig.3 . 
The FLOPs count of the D3 block significantly dominates the total FLOPs count. This unbalanced FLOPs distribution is quite different from that of the ResNet architecture, which has small variant across different blocks. The unbalanced FLOPs distribution gives us the chance to get a good overall FLOPs reduction percentile by only reducing the cost of the heaviest layer. Quantitatively, for example, if we can reduce the FLOPs of D3 by half, the total FLOPs can be reduced by 24\%. In the following subsections, we will discuss the main insights of how to get a tiny back-end.

\subsection{Tiny back-end solution}
\subsubsection{Light-weight block}
Intuitively, we can reduce the filter size in order to get FLOPs reduction. As shown in Fig. 4, we have propose different block design for the detection branches of ResNet. The D-block-v1 is applying the MobileNet \cite{mobilenet} building block here. A 3x3 depth-wise (dw) convolution is followed by a 1x1 convolutional block to substitute one orginal layer. The D-block-v2 alternately places 1x1 and 3x3 kernels. This one is inspired by the YOLOv1 \cite{yolov1}, which has replaces the 3x3 kernels without introducing residual blocks. In our design, we even make it simpler to keep number of filters fixed across different layers. The D-block-v3 is more aggressive, which replaces all the 3x3 convolutions with 1x1 convolutions. 

\begin{figure*}[!t]
\begin{center}
\includegraphics[width=8cm]{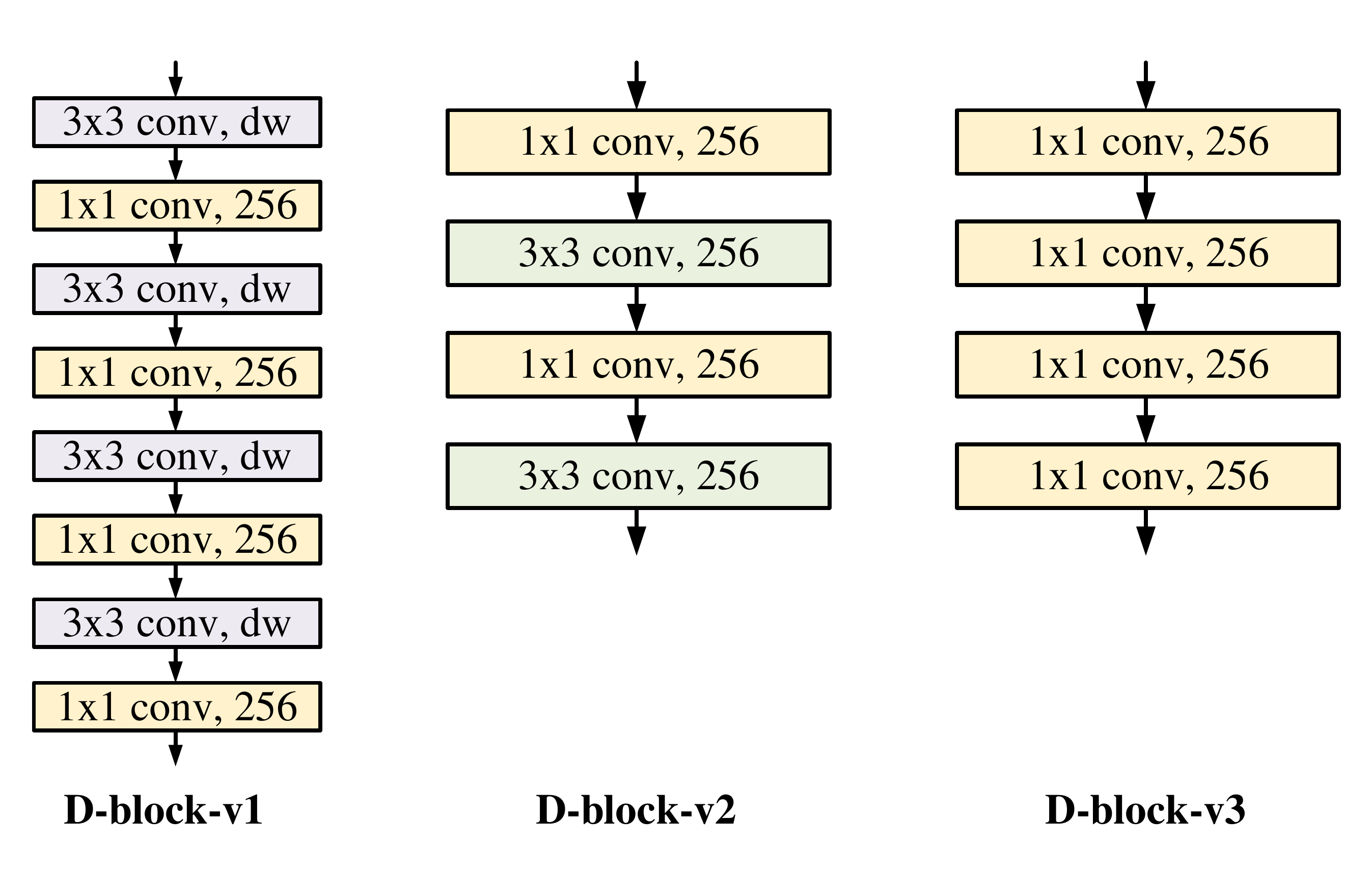}
\end{center}
   \caption{light-weight blocks for detection backend.}
\label{fig:short}
\end{figure*}

Apparently, the light-weight block will cause certain accuracy drop to tradeoff less computation cost. Therefore, we introduce limited overheads to compensate the accuracy drop here, which is stated in the next subsection.

\subsubsection{Partially shared weights}
As illustrated in Section 3.2.1., the light-weight detection blocks are trading off lower computational complexity with accuracy drop. To compensate the accuracy drop, we proposed to replace the fully shared weight scheme in the original RetinaNet to a partial shard weight scheme. As shown in Fig. 3, P3-P7 is the multi-scale feature maps outputs of FPN, which then feed into detection backend D3-D7 respectively. Although D3-D7 shared the weight parameters, D3-D7 have unique input size (P3-P7) and are processed in serial. Fig. 3(a) is the original one that D3-D7 fully share the weights. In Fig. 3(b), only D4-D7 share the weights with original configuration, while D3 will be processed by the light-weight D-block-v1/v2/v3 that proposed in Section 3.2.1. 

\begin{figure*}[!t]
\begin{center}
\includegraphics[width=11cm]{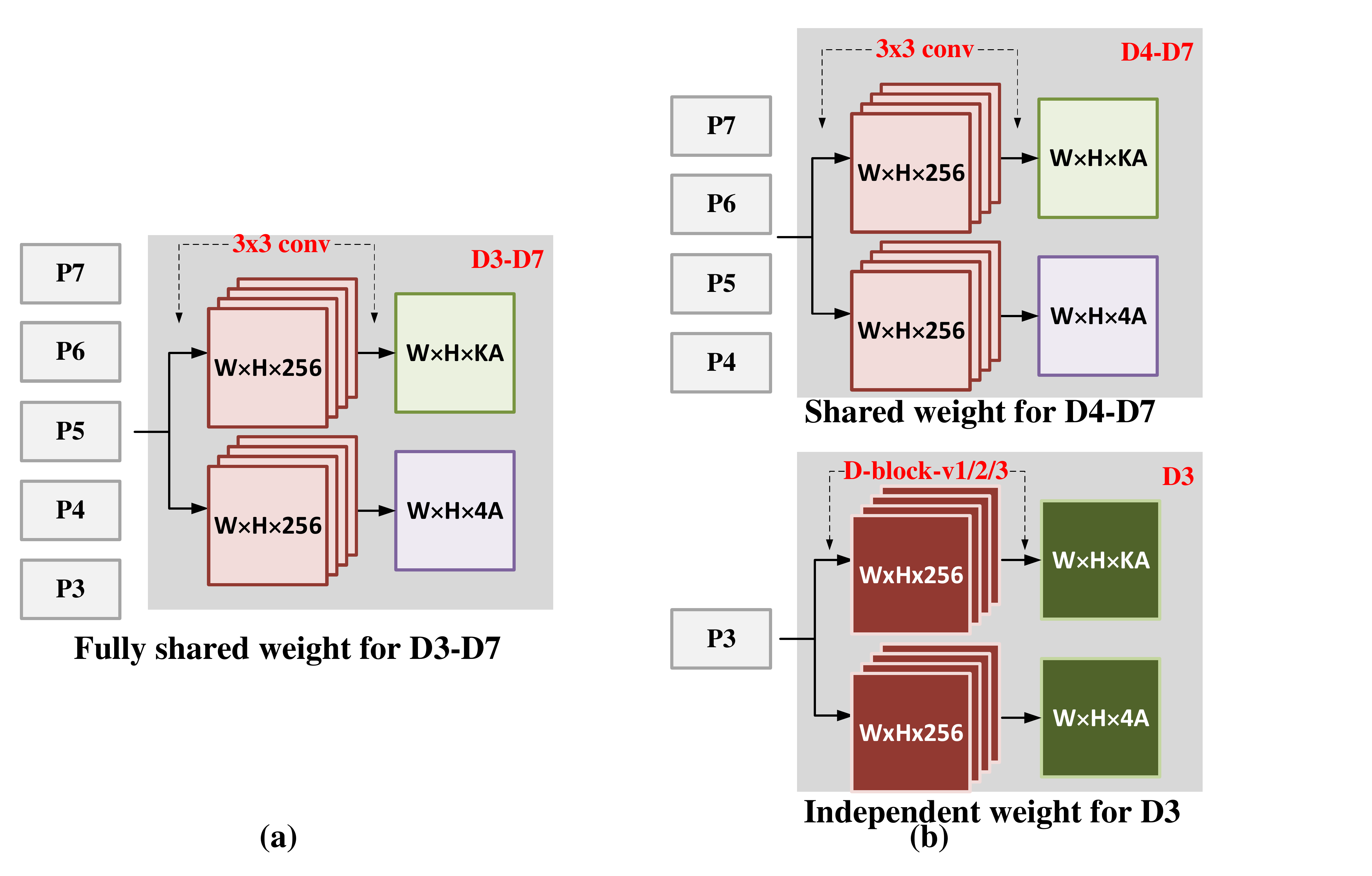}
\end{center}
   \caption{Fully and partially shared weights for detection backend.}
\label{fig:short}
\end{figure*}

Partially shared weights scheme mainly has two advantages. For one thing, as D3 has its independent weight parameters, it can learn more tailored features for its branch which can compensate the accuracy drop brought by lower computational complexity. For another, this enables us not to touch the rest of the network but only solving the heaviest bottleneck block. Also, as the backbone (ResNet-50) dominate the memory consumption (as shown in Fig. 3), the overhead of memory consumption here can be negligible. Quantitatively, weight parameter increment introduced here is less than 1\% of the total weights. 

\section{Results and discussion}
\subsection{Experimental setup}
We perform our experiments in Caffe2 with 4 Titan X GPUs. We build upon the open source code of RetinaNet in \cite{Detectron2018}. As the original work is training with 8 GPUs, we scales down the base learning rate by 2x and extend the training epochs by 2x as suggested in \cite{accuratesgd}. Besides, \cite{rethinkingpruning} proves the deep neural network is less easier to overfit when its computational complextity is reduced by network compression. It also suggests to extend the training time accordingly for better accuracy rate. As we reduce the total FLOPs in light-weight RetinaNet, we further extend the training epoch with the same factor of FLOPs reduction. In all the experiments, we fix the network configuration as RetinaNet-ResNet50-FPN.

\subsection{Performance on COCO dataset}
The COCO dataset \cite{cocodataset} is the considered as the most complicated one for object detection. As we are targeting on discussing the trade-off for high-end object detection network, we only perform experiments on COCO dataset (same as the original RetinaNet does \cite{retinanet}). We train the light-weight RetinaNet on 2017 COCO training dataset and test it on COCO test-dev.

Table 1 shows the comparison among different light-weight blocks that we proposed in Section 3.2.1. In this set of experiment, we only use the light-weight block in the regression branch (for the bounding box) of detection backend, which is the upper branch shown in Fig. 2 detection backend. The results of Table 1 show that the D-block-v1 -- the one with the MobileNet building block has 0.8\% lower mAP compared with the D-block-v3, which has the same FLOPs reduction percentile. It also aligns with our analysis in Section 3.2 that although MobileNet is proved to a powerful light-weight classification network architecture, MobileNet building block is not guaranteed to be the best building block substitution for other vision-based tasks. Therefore, with the same scale of FLOPs reduction, we will choose D-block-v3 instead of D-block-v1 in the following experiment. As the D-block-v2 performs less aggressive FLOPs reduction, its mAP is only reduced by 0.1\%, which is a good trade-off for small scale FLOPs reduction (15\%).

\begin{table}[!t]
\begin{center}
\includegraphics[width=9cm]{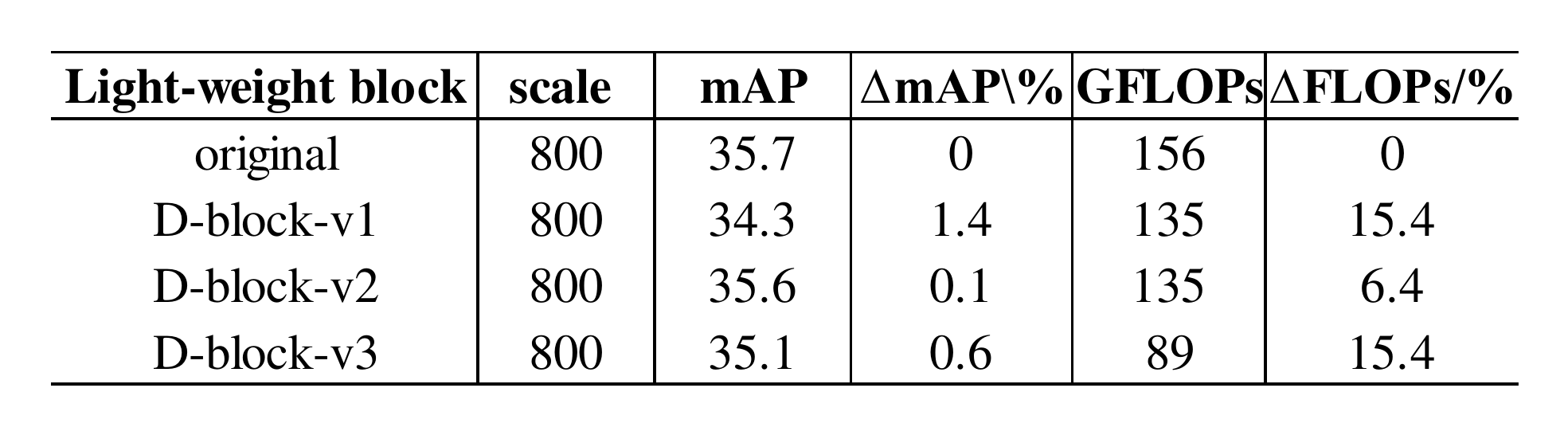}
\vspace*{-8mm}
\end{center}
\caption{Comparison between different light-weight block.}
\end{table}

The configurations for different versions of light-weight RetinaNet with D-block-v2 or D-block-v3 light-weight blocks are shown in Table 2. Table 2 shows the information of which light-weight block is applied to one or both of classification and bounding box (regression) branch. The corresponding light-weight RetinaNet performance results are shown in Table 3. As scaling down input image size is the most common practice for FLOPs and accuracy trade-off, we also list the performance results of original RetinaNet with different input scales (we directly cited the results from the RetinaNet paper). To give a more straightforward understanding of the proposed method versus input image scaling, we visualize the FLOPs and accuracy trade-off in Fig. 6. Each data point in Fig. 6 is corresponding to one row of results in Table 3. We mark the trending line of light-weight RetinaNet in red dot line and that of original RetinaNet in blue dot line. As we stated before, the upper-left corner means the best trade-off in this kind of mAP-FLOPs graph. As the red line is constantly closer to the upper-left corner, it indicates that the proposed method has better mAP-FLOPs trade-off than the conventional input image scaling method. The difference between these two methods result in 0.1\% mAP at the same number of FLOPs with low reduction ratio. However, as we further reduce the number of FLOPs, the proposed method shows a trend in linear degradation, while the input image scaling method degrade in a more exponentially direction. Fig. 6 clearly shows a divergence at the GFLOPs around 90, where the conventional method is yielded to 0.3\% more accuracy drop than the proposed method.

\begin{table}[!t]
\begin{center}
\includegraphics[width=10cm]{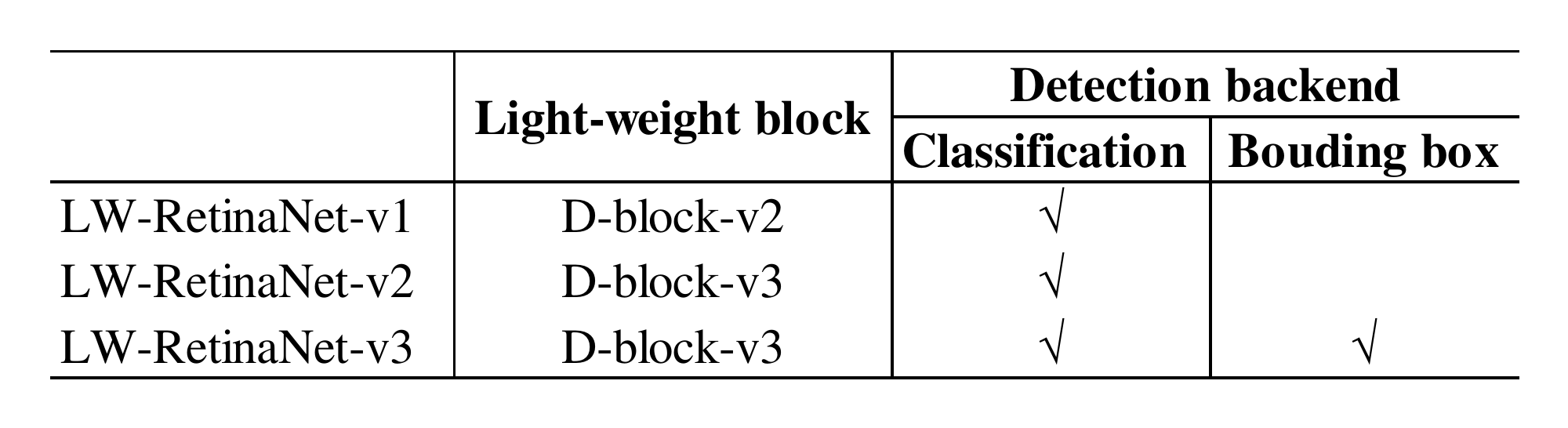}
\end{center}
\vspace*{-8mm}
\caption{Configurations of different light-weight(LW) RetinaNet.}
\end{table}

\begin{table}[!t]
\begin{center}
\includegraphics[width=12.5cm]{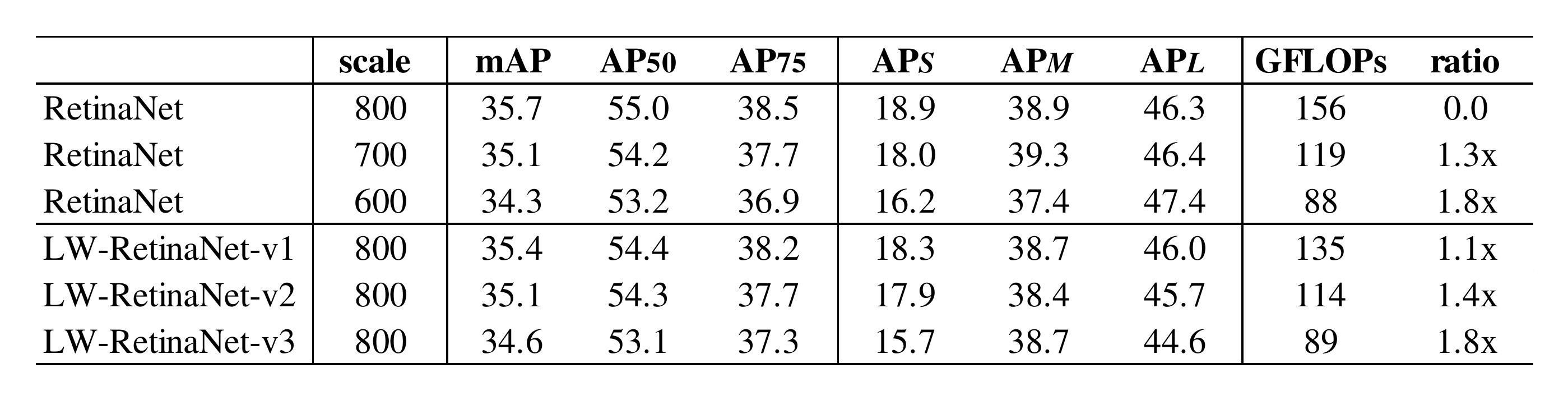}
\end{center}
\vspace*{-8mm}
\caption{Comparison of original RetinaNet and proposed light-weight RetinaNet.}
\end{table}

\begin{figure*}[!t]
\begin{center}
\includegraphics[width=10cm]{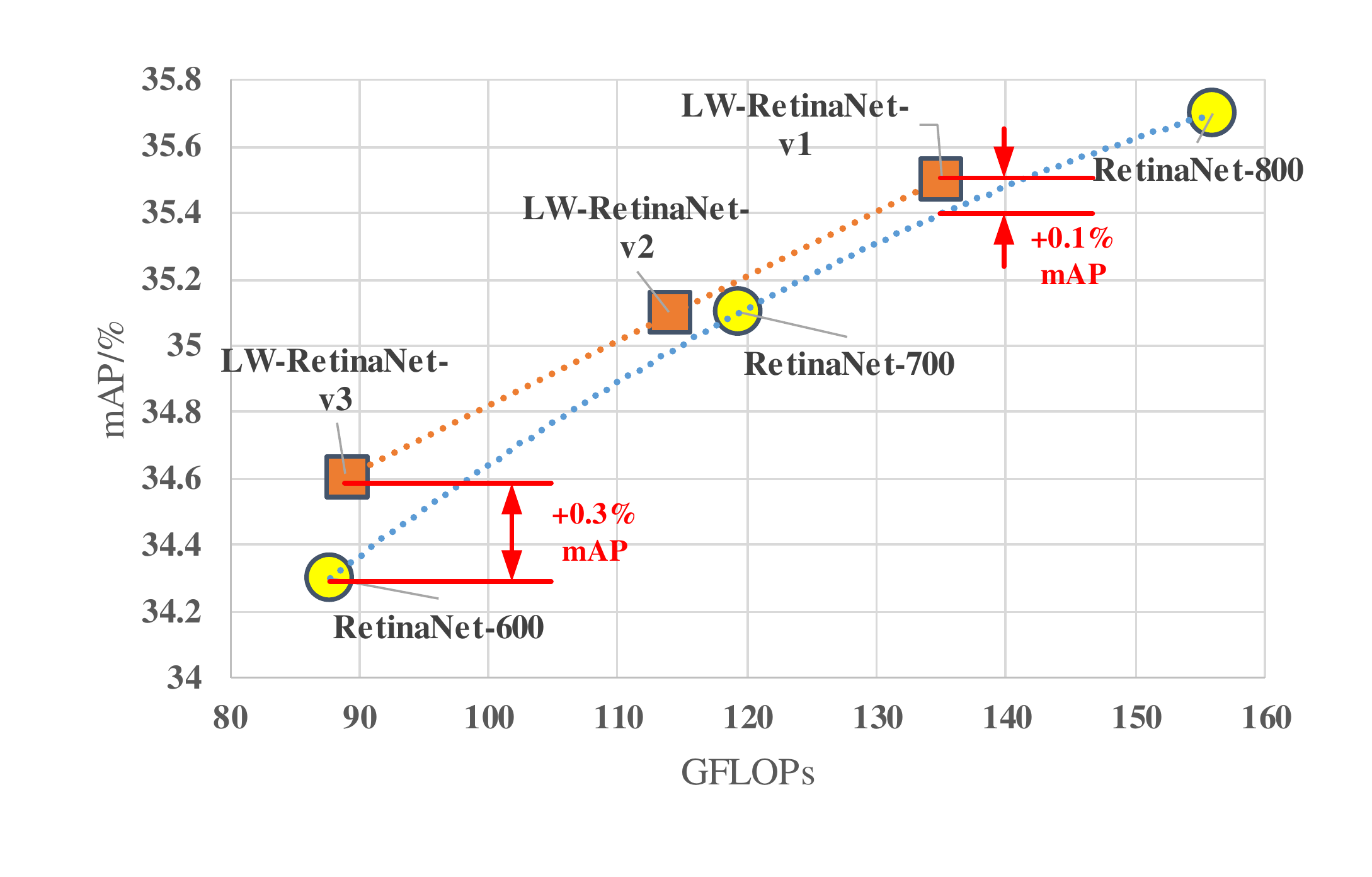}
\vspace*{-8mm}
\end{center}
   \caption{FLOPs and mAP trade-off for input image size scaling versus the proposed method.}
\label{fig:short}
\end{figure*}
As any detection methods with FPN structure can result in an imbalanced FLOPs distribution, the proposed method can be potentially applied to such kinds of detection network for a better choice for mAP-FLOP.

\section{Conclusion}
In this paper, We proposed only to reduce the FLOPs in the heaviest bottleneck layer for a blockwise-FLOPs-imbalance RetinaNet to get its light-weight version. The proposed solution shows a constantly better mAP-FLOPs trade-off line in a linear degradation trend, while the input image scaling method degrades in a more exponentially trend. Quantitatively, the proposed method result in 0.1\% mAP improvement at 1.15x FLOPs reduction and 0.3\% mAP improvement at 1.8x FLOPs reduction. The proposed method can be potentially applied to any FPN-based blockwise-FLOPs-imbalance detection network.



\bibliography{egbib}
\end{document}